\documentclass{article}

\usepackage[final]{preprint-sub}
\usepackage[usestackEOL]{stackengine}

\usepackage[utf8]{inputenc} 
\usepackage[T1]{fontenc}    
\usepackage{hyperref}       
\usepackage{url}            
\usepackage{booktabs}       
\usepackage{amsfonts}       
\usepackage{nicefrac}       
\usepackage{microtype}      
\usepackage{xcolor}         

\title{LSTM based models stability in the context of Sentiment Analysis for social media}

\author{
  Bousselham EL HADDAOUI, Raddouane CHIHEB, Rdouan FAIZI, and Abdellatif EL AFIA\\
  ENSIAS, MOHAMMED V UNIVERSITY IN RABAT, MOROCCO\\
  \texttt{haddaoui.bousselham@gmail.com,  r.chiheb@um5s.net.ma,} \\
  \texttt{r.faizi@um5s.net.ma, a.elafia@um5s.net.ma} \\
}

\begin{document}

\maketitle

\begin{abstract}
Deep learning techniques have proven their effectiveness for Sentiment Analysis (SA) related tasks. Recurrent neural networks (RNN), especially Long Short-Term Memory (LSTM) and Bidirectional LSTM, have become a reference for building accurate predictive models. However, the models complexity and the number of hyperparameters to configure raises several questions related to their stability. In this paper, we present various LSTM models and their key parameters, and we perform experiments to test the stability of these models in the context of Sentiment Analysis.
\end{abstract}

\section{Introduction}
Sentiment Analysis is a research field that focuses on extracting opinions and sentiments from social media and other sources in an automated way. It's also seen as a complex opinion mining problem [1], this complexity has been addressed in various aspects through previous research resulting to a unified framework with several components such as data acquisition, preprocessing, classification and visualization [2]. Machine learning and deep learning for classification, combined with data representation techniques such as word embedding, are state-of-the-art stack for a SA solution. However, this combination results in a high number of hyperparameters and model complexity.
	 	 	 		
In the context of social media, the content is usually a short text with few sentences. The sequential aspect of words in these sentences make it a suitable use case for classification using recurrent neural networks (RNNs) [3]. RNN architecture suffers from a known vanishing gradient problem which is linked to error vanishing as it propagates back [4], this issue is addressed in its variants LSTM [5] and BiLSTM [6]. 
Neural networks models deal with numerical inputs, Word Embeddings [7] were introduced as a new approach that maps textual information to a numeric form and encodes syntactic and semantic information in denser feature spaces. They can be categorized into two types [8]: prediction-based models which are context aware and leverage local data, and count-based models that tend to use global information like word count, frequencies, etc. The most frequently used word embeddings are Word2vec [9] and Global Vectors (GloVe) [10].

In our study, we made experiments on various LSTM and BiLSTM architectures to test the stability of each model against its hyperparameters variations and the provided word embeddings. The rest of this paper is organized as follows. Section 2 provides a detailed description of the system. The experimentation process and the results are discussed in section 3. Conclusions are summarized in section 4.

\section{System description}
The system’s core model is designed as stacked layers of LSTM and BiLSTM neural networks. The hyperparameters investigated in the study are: the learning rate, the optimizer, the network size, the network depth, the batch size, the number of epochs, the weights initialization, the loss function and the dropout.
The dataset used for the experiments was constructed from a dataset used in the SemEval competition for sentiment analysis between the edition 2013 and 2016 [11], [12], [13], [14]. 50 133 tweets in English language gathered for our analysis with the following ratios: 44.9\% neural, 39.5\% positive and 15.6\% negative.
		
Since text preprocessing isn’t the main scope of our study, we  relied only on a few recommended [15] tasks such as: removing stop words, URLS, mentions and hashtags. For word representation transformation, we used a general purpose pre-trained word vectors using GloVe  from a 2 billion tweets corpus. The pre-trained word vectors are intended to be twitter general purpose and not domain specific, and they are provided with various vector dimensions 25,50,100 and 200. The learned embedding matrix from the training dataset and word embedding’s initialize the embedding layer in our models. In addition to Glove, we use also Google News pre-trained vectors (Word2vec with 300 vector dimension) to further investigate the stability of our models to the word embedding category.

\section{Experiment and results}

\begin{table}[hbt!]
  \caption{Experiments results and observations}
  \label{sample-table}
  \centering
  \begin{tabular}{lll}
    \toprule
    Test     & Results     & Observations  \\
    \midrule
    T1 & 0.8451  & 
	\Longstack[l]{ - Model: BiLSTM2, batch size=500, epochs=200, GloVe with dimension=100,\\ dropout=20\%, layers hidden units=100 and optimizer=Adam \\
	- An unstable behavior when using deep architectures with low epochs }\\
    \midrule
    T2 & 0.8461 & 
	\Longstack[l]{ - Model: LSTM1, dropout= 25\%, GloVe with dimension=100, batch size=500,\\ epochs=200,layers hidden units=100 and optimizer=Adam \\
	- The dropout can improve the model’s performance, but it affects its stability\\ when changing size and depth }\\
    \midrule
    T3 & 0.8431 & 
	\Longstack[l]{ - Model: BiLSTM3, GloVe with dimension=100, batch size=500, dropout=25\%,\\ epochs=200, layers hidden units=100 and optimizer=Adam \\
	- Models tend to have a stable behavior when using high dimensions\\ (i.e., 100 or 200) }\\
    \midrule
    T4 & 0.8729 & 
	\Longstack[l]{ - Model: BiLSTM1 with ADAM [15] optimizer, GloVe with dimension=100,\\ batch size=500, dropout=25\%, epochs=200, layers hidden units=100 \\
	- Adam gives a stable performance with variant model depths comparing\\ to SGD [14] and RMSPROP }\\
    \midrule
    T5 & 0.8743 & 
	\Longstack[l]{ - Model: LSTM1, GloVe with dimension 200, batch size=500, dropout=25\%,\\ epochs=200, layers hidden units=100 and optimizer=Adam \\
	- Models are stable with both GloVe and Word2vec, but performing \\slightly better with GloVe }\\
    \bottomrule
  \end{tabular}
\end{table}

The models are divided into two categories; LSTM and BiLSTM models, each category providing three variants with respectively 1,2 and 3 layers. We name our models for the rest of the paper as LSTM1, LSTM2, LSTM3, BiLSTM1, BiLSTM2 and BiLSTM3 denoting the model type and the number of used layers.
The performed experiments are T1: The sensitivity of LSTM and BiLSTM to batch size and epochs, T2: The sensitivity of LSTM and BiLSTM to dropout, T3: The sensitivity of LSTM and BiLSTM to word embeddings dimensions, T4: The sensitivity of LSTM and BiLSTM to optimizers, and T5: The sensitivity of LSTM and BiLSTM to word embeddings type. 

The evaluation process is performed on validation data with a ratio of 20\% the size of the dataset, the monitored metric to compare models is the best validation accuracy. Rresults to our experiments are presented in Table 1 showcasing the best achieved performance and major observations for each test case:

\section{Conclusion}
In this paper we studied the stability of LSTM based models to their hyperparameters and word embeddings techniques. Experiments results shows that certain hyperparameters values changes don’t affect LSTM based models stability such as word embeddings category, some optimizers (i.e., ADAM) and high word embeddings dimensions. Whereas network depth, size and dropout seem to directly affect the models stability. Future work will investigate the effect of domain trained word embeddings and the learning rate on our models.

\section*{References}

\medskip

{
\small

[1]	B. Liu, "Sentiment Analysis and Opinion Mining,"  Synth. Lect. Hum. Lang. Technol., vol. 5, no. 1, pp. 1–167, May 2012, doi: 10.2200/S00416ED1V01Y201204HLT016.

[2]	B. El Haddaoui, R. Chiheb, R. Faizi, and A. El Afia, “Toward a Sentiment Analysis Framework for Social Media,” 2018, pp. 1–6. doi: 10.1145/3230905.3230919.

[3]	W. Yin, K. Kann, M. Yu, and H. Schütze, “Comparative Study of CNN and RNN for Natural Language Processing,” ArXiv170201923 Cs, Feb. 2017, Accessed: Nov. 14, 2020. [Online]. Available: http://arxiv.org/abs/1702.01923

[4]	S. Hochreiter, “The Vanishing Gradient Problem During Learning Recurrent Neural Nets and Problem Solutions,” Int. J. Uncertain. Fuzziness Knowl.-Based Syst., vol. 06, no. 02, pp. 107–116, Apr. 1998, doi: 10.1142/S0218488598000094.

[5]	S. Hochreiter and J. Schmidhuber, “Long Short-Term Memory,” Neural Comput., vol. 9, no. 8, pp. 1735–1780, Nov. 1997, doi: 10.1162/neco.1997.9.8.1735.

[6]	M. Schuster and K. K. Paliwal, "Bidirectional recurrent neural networks,"IEEE Trans. Signal Process., vol. 45, no. 11, pp. 2673–2681, Nov. 1997, doi: 10.1109/78.650093.

[7]	J. Turian, L.-A. Ratinov, and Y. Bengio, “Word Representations: A Simple and General Method for Semi-Supervised Learning,” p. 11, 2010.

[8]	F. Almeida and G. Xexéo, “Word Embeddings: A Survey,” ArXiv190109069 Cs Stat, Jan. 2019, Accessed: Nov. 21, 2020. [Online]. Available: http://arxiv.org/abs/1901.09069

[9]	T. Mikolov, K. Chen, G. Corrado, and J. Dean, “Efficient Estimation of Word Representations in Vector Space,” ArXiv13013781 Cs, Sep. 2013, Accessed: Nov. 14, 2020. [Online]. Available: http://arxiv.org/abs/1301.3781

[10]	J. Pennington, R. Socher, and C. Manning, “Glove: Global Vectors for Word Representation,” in Proceedings of the 2014 Conference on Empirical Methods in Natural Language Processing (EMNLP), Doha, Qatar, 2014, pp. 1532–1543. doi: 10.3115/v1/D14-1162.

[11]	J. Y. Lee and F. Dernoncourt, “Sequential Short-Text Classification with Recurrent and Convolutional Neural Networks,” in Proceedings of the 2016 Conference of the North American Chapter of the Association for Computational Linguistics: Human Language Technologies, San Diego, California, 2016, pp. 515–520. doi: 10.18653/v1/N16-1062.

[12]	P. Nakov, S. Rosenthal, Z. Kozareva, V. Stoyanov, A. Ritter, and T. Wilson, “SemEval-2013 Task 2: Sentiment Analysis in Twitter,” in Second Joint Conference on Lexical and Computational Semantics (*SEM), Volume 2: Proceedings of the Seventh International Workshop on Semantic Evaluation (SemEval 2013), Atlanta, Georgia, USA, Jun. 2013, pp. 312–320. Accessed: Nov. 23, 2020. [Online]. Available: https://www.aclweb.org/anthology/S13-2052

[13]	S. Rosenthal, A. Ritter, P. Nakov, and V. Stoyanov, “SemEval-2014 Task 9: Sentiment Analysis in Twitter,” in Proceedings of the 8th International Workshop on Semantic Evaluation (SemEval 2014), Dublin, Ireland, Aug. 2014, pp. 73–80. doi: 10.3115/v1/S14-2009.

[14]	S. Rosenthal, S. M. Mohammad, P. Nakov, A. Ritter, S. Kiritchenko, and V. Stoyanov, "SemEval-2015 Task 10: Sentiment Analysis in Twitter," ArXiv191202387 Cs, Dec. 2019, Accessed: Nov. 23, 2020. [Online]. Available: http://arxiv.org/abs/1912.02387

[15]	D. Effrosynidis, S. Symeonidis, and A. Arampatzis, “A Comparison of Pre-processing Techniques for Twitter Sentiment Analysis,” in Research and Advanced Technology for Digital Libraries, vol. 10450, J. Kamps, G. Tsakonas, Y. Manolopoulos, L. Iliadis, and I. Karydis, Eds. Cham: Springer International Publishing, 2017, pp. 394–406.

}

\end{document}